\def\year{2021}\relax
\documentclass[letterpaper]{article} 
\usepackage{aaai21}  
\usepackage{times}  
\usepackage{helvet} 
\usepackage{courier}  
\usepackage[hyphens]{url}  
\usepackage{graphicx} 
\usepackage{natbib}  
\usepackage{caption} 
\usepackage{amsmath}
\usepackage{amssymb}
\usepackage[ruled]{algorithm2e}
\usepackage{multirow}
\usepackage{subfigure}
\usepackage{bm}
\usepackage[switch]{lineno}

\DeclareMathOperator{\Tr}{Tr}

\title{Deep Semantic Dictionary Learning for Multi-label Image Classification}
\author{
   Fengtao Zhou\textsuperscript{$1$},
   Sheng Huang\textsuperscript{$1$,$2$,\footnote{indicates the corresponding author.}},
   Yun Xing\textsuperscript{$1$}\\
}
\affiliations{
   $^1$School of Big data \& Software Engineering, Chongqing University,\\
  $^2$Ministry of Education Key Laboratory of Dependable Service Computing in Cyber Physical Society,\\
   Shazheng street NO.174, Shapingba District, Chongqing 400044, China\\
   \{zft, huangsheng, yxing\}@cqu.edu.cn
}
\begin{document}
\maketitle

\begin{abstract}
    Compared with single-label image classification, multi-label image classification is more practical and challenging. Some recent studies attempted to leverage the semantic information of categories for improving multi-label image classification performance. However, these semantic-based methods only take semantic information as type of complements for visual representation without further exploitation. In this paper, we present an innovative path towards the solution of the multi-label image classification which considers it as a dictionary learning task. A novel end-to-end model named Deep Semantic Dictionary Learning (DSDL) is designed. In DSDL, an auto-encoder is applied to generate the semantic dictionary from class-level semantics and then such dictionary is utilized for representing the visual features extracted by Convolutional Neural Network (CNN) with label embeddings. The DSDL provides a simple but elegant way to exploit and reconcile the label, semantic and visual spaces simultaneously via conducting the dictionary learning among them. Moreover, inspired by iterative optimization of traditional dictionary learning, we further devise a novel training strategy named Alternately Parameters Update Strategy (APUS) for optimizing DSDL, which alternately optimizes the representation coefficients and the semantic dictionary in forward and backward propagation. Extensive experimental results on three popular benchmarks demonstrate that our method achieves promising performances in comparison with the state-of-the-arts. Our codes and models have been released\footnote{https://github.com/ZFT-CQU/DSDL.}.
\end{abstract}

\section{Introduction}
\begin{figure}[t]
    \centering
    \includegraphics[scale=0.55]{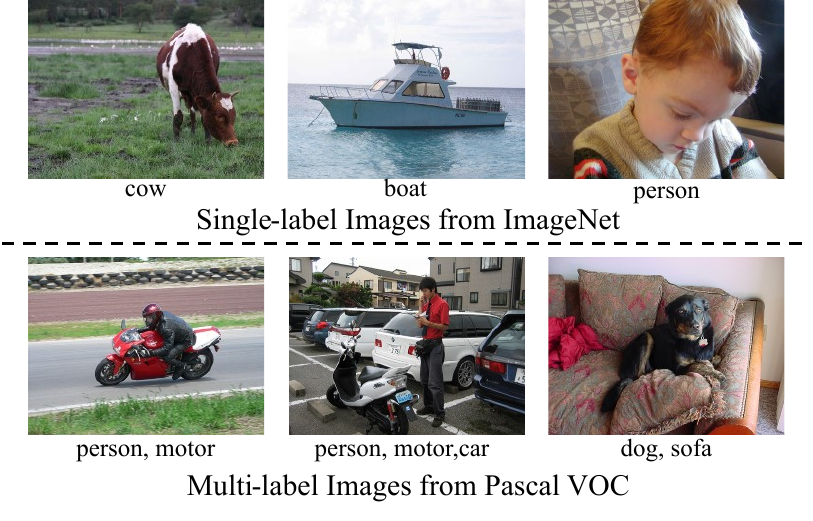}
    \caption{Some examples from ImageNet and Pascal VOC. The foreground objects in single-label images are roughly aligned. However, the assumption of object alignment is not valid for multi-label images in which different objects are located at various positions under different scales and poses.}
    \label{fig1}
\end{figure}

With the emergence of large-scale datasets, \textit{e.g.} ImageNet, Convolutional Neural Networks (CNNs) have achieved great success for single-label image classification due to their powerful representation learning abilities. However, most of real-world images contain more than one category of objects which are located at various positions and scales with complex background. As depicted in Figure~\ref{fig1}, the foreground objects of single-label images are roughly aligned while this assumption is usually invalid for multi-label images. Furthermore, the different composition and interaction between objects in multi-label images also increases the complexity of multi-label image classification. In such a manner, the multi-label image classification is more practical and challenging than single-label image classification.

To tackle the multi-label classification problem, extensive approaches have been proposed over the past decades. Generally speaking, they can be divided into two categories~\cite{zhang2013review}, \textit{i.e.}, problem transformation methods and algorithm adaptation methods. The problem transformation methods solve the multi-label image classification task by transforming it into other well-established classification scenarios. The most straightforward way is to treat multi-label image classification as a set of binary classification, and to train the independent classifier for each class with cross-entropy~\cite{tagprop} or ranking loss~\cite{gong2013deep}. The algorithm adaption methods address the multi-label image classification problem via adapting popular learning techniques to process multi-label images directly, such as modifying the output layer of single-label classification model~\cite{he2016deep, simonyan2014very}. Briefly, the philosophy of problem transformation methods is to fit data to algorithm, while the algorithm adaptation is to fit algorithm to data.

Influenced by the recent advancement in Neural Language Processing (NLP)~\cite{word2vec}, the role of semantic information becomes increasingly important in many fields, such as zero-shot learning~\cite{fgn, cpdn}, few-shot learning~\cite{acm}, image annotation~\cite{aia} and fine-grained categorization~\cite{akata2015evaluation, he2017fine}. In the last few years, some researchers attempted to leverage the semantic information of categories to improve multi-label image classification performance~\cite{chen2019multi,chen2019learning,mlzsl}. Nevertheless, these works only consider the semantic information as the complements for visual features or the side-information for guiding the derivation of the discriminative classifiers, while no one focus on exploiting and analyzing the correlations among label, semantic and visual spaces, which may provide a potential way for further improving the multi-label image classification. Dictionary learning has been proven as an effective tool for exploiting the correlation between label and visual spaces in various fields, such as image annotation~\cite{mhdsc}, multi-label embedding~\cite{cdl} and multi-instance learning~\cite{maxmargin}. However, all these relevant works are based on shallow learning paradigm associated with multiple linear transformations and dedicated optimization algorithm which limits the effectiveness and flexibility of the model. Moreover, the advanced semantic representation, \textit{e.g.} word embedding, has not been incorporated yet, and the correlation between semantic and visual spaces still remains unstudied.

In this paper, we present a novel end-to-end multi-label image classification approach named Deep Semantic Dictionary Learning (DSDL) to address the aforementioned issues. DSDL considers the multi-label image classification problem as a dictionary learning task for solution. It leverages an auto-encoder to generate the semantic dictionary aligned with the visual space from class-level semantics, so the visual features extracted by CNN can be represented by semantic dictionary with their label embeddings. Unlike conventional multi-label image classification approaches, DSDL not only captures the correlations between the label and visual spaces, but also reconciles the relations of the label, semantic and visual spaces. The contributions of our work can be summarized as follows,
\begin{itemize}
    \item We present a succinct but effective approach named Deep Semantic Dictionary Learning (DSDL) for multi-label image classification, which can exploit the label, semantic and visual spaces simultaneously. To the best of our knowledge, it is the first deep dictionary learning introduced to the multi-label image classification field.
    \item We design a novel training strategy, Alternately Parameter Updating Strategy (APUS), for Deep Semantic Dictionary Learning, which updates representation coefficients and semantic dictionary alternately in the forward and backward propagation.
    \item We evaluate our model on three multi-label image benchmarks, Pascal VOC 2007, 2012 and Microsoft COCO. Extensive experimental results demonstrate the proposed method achieves promising performances in comparison with the state-of-the-arts.
\end{itemize}

\section{Related work}
\subsection{Multi-label Image Classification}
\subsubsection{Problem Transformation Methods}
In this kind of approaches, the straightforward way is to treat multi-label image classification as a set of binary classification, and to train the independent classifier for each class with cross-entropy~\cite{tagprop} or ranking loss~\cite{gong2013deep}. Clearly, these approaches neglect the label dependencies while the images with multiple objects contain strong correlation among labels in nature~\cite{rlsd}. Conventionally, the problem transformation approaches often leverage probabilistic graphical models to address this issue~\cite{cg,xue2011correlative,guo2011multi}. They model the co-occurrence dependencies with pairwise compatibility probabilities or co-occurrence probabilities, and then utilize the graphical models, such as Markov random fields~\cite{guo2011multi} and conditional random fields~\cite{ltcrf}, to infer the final joint label probabilities. However, for these above-mentioned methods, a large number of sub-classifiers (or sub-models) have to be trained for a multi-label classification task which obviously reduces the efficiency and practicability of model, and inevitably induces parameters redundancy problem.

\subsubsection{Algorithm Adaptation Methods}
In recent years, algorithm adaptation methods are the mainstream method for multi-label image classification, since it provides a more natural and intuitive solution. Inspired by remarkable successes of deep learning in the recent decades, increasing people devoted to developing deep learning-based methods to address multi-label image classification because they are very flexible to adapt single-label classification into multi-label classification such as simply adjusting the output layer~\cite{he2016deep, simonyan2014very}. Most recently, Wang et al~\cite{wang2016cnn} presented a unified CNN-RNN framework which employs CNN and RNN to model the label co-occurrence dependency in a joint image/label embedding space for multi-label image classification. Chen et al~\cite{chen2018recurrent} utilized Long-Short Term Memory (LSTM) to develop the recurrent attention reinforcement learning framework which is capable of automatically discovering the attentional and informative regions related to different semantic objects and further predict label scores conditioned on these regions. Zhu et al~\cite{zhu2017learning} proposed a spatial regularization network to capture both semantic and spatial relations of these multiple labels based on weighted attention maps. Motivated by the recent success of Graph Convolutional Network (GCN), Chen et al~\cite{chen2019multi} leveraged GCN for learning the label classifiers and modeling label dependencies simultaneously.

\subsection{Dictionary Learning}
Previously, dictionary learning has been proven as an effective tool for exploiting the correlation between label and visual spaces in the field of image annotation, multi-label embedding and multi-instance learning. For example, Jing et al~\cite{jing2016multi} proposed Multi-Label Dictionary Learning (MLDL) model, which can conduct multi-label dictionary learning in input feature space and partial-identical label embedding in output label space simultaneously. Cao et al~\cite{cao2015sled} presented the Semantic Label Embedding Dictionary (SLED) model for image annotation under a weakly supervised setting, which can explicitly fuse the label information into dictionary representation and explore the semantic correlations between co-occurrence labels. Li et al~\cite{mvmil} developed a novel multi-view dictionary learning algorithm to learn a multi-view graph dictionary which considers cues from all views to improve the discrimination of the multi-instance learning model.

Different from all these methods above mentioned, our proposed approach can exploit and reconcile the label, semantic and visual spaces simultaneously via representing visual features by semantic dictionary with label embeddings \cite{liu2009joint}.

\begin{figure*}[t]
    \centering
    \includegraphics[scale=0.78]{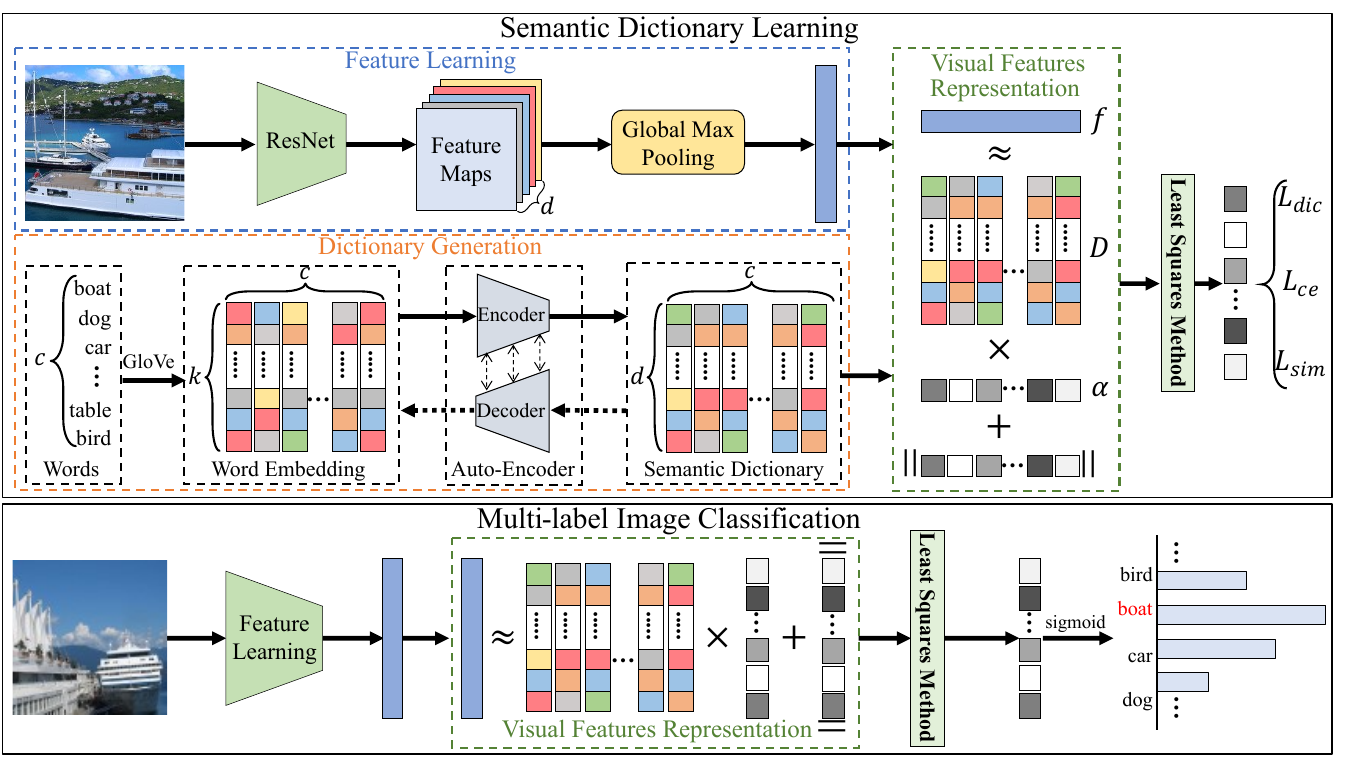}
    \caption{Illustration of Deep Semantic Dictionary Learning. The semantic dictionary learning stage (training stage) aims at training CNN for feature extraction and auto-encoder for semantic dictionary generation. In the multi-label image classification stage (testing stage), the label occurrence probabilities of given image can be estimated by encoding its visual features with the learned semantic dictionary. }
    \label{overview_pic}
    \vspace{-0.2cm}
\end{figure*}

\section{Methodology}
\subsection{Problem Formulation and Notation}
Suppose $\mathcal{X}\triangleq \mathcal{R}^{w^\prime \times h^\prime \times c^\prime }$ denotes the instance space and $\mathcal{Y}\triangleq \{y^1, y^2, \dots ,y^c \}$ denotes the label space with $c$ possible classes where $y^i$ is 1 if the class is relevant and 0 otherwise. The task of multi-label image classification aims to learn a multi-label image classifier $\mathcal{T}(\cdot)$ trained on the training set $\mathcal{D}=\{(x_i,y_i)_{i=1}^N\}$,
\begin{equation}
    \mathcal{X}\stackrel{\mathcal{T}}{\longrightarrow} 2^\mathcal{Y} \label{ProblemFormulation}.
\end{equation}
For each multi-label image instance $(x_i,y_i)$, $x_i \in \mathcal{X}$ is a real-world image and $y_i \in \mathcal{Y}$ is the set of labels associated with $x_i$. In our proposed model, we decompose $\mathcal{T}(\cdot)$ into $\mathcal{T}(\cdot) =\mathcal{J}(\mathcal{F}(\cdot))$ where $\mathcal{F}(\cdot):\mathcal{R}^{w^\prime \times h^\prime  \times c^\prime }\rightarrow \mathcal{R}^d$ is the feature learning module that extracts $d$-dimensional features from original image, and $\mathcal{J}(\cdot):\mathcal{R}^d\rightarrow \mathcal{R}^c$ is the feature representation module that encodes visual features into the $c$-dimensional label space $\mathcal{Y}$ by semantic dictionary. In the testing stage, for any sample without annotation $x\in \mathcal{X}$, the multi-label classifier $\mathcal{T} (\cdot)$ predicts $\mathcal{T}(x)\subseteq \mathcal{Y}$ as the set of proper labels for $x$,
\begin{equation}
    \hat{y} = \mathcal{T}(x) = \mathcal{J}(\mathcal{F}(x)).
    \label{classifier}
\end{equation}

\subsection{Overview}
We present a novel multi-label image classification method named Deep Semantic Dictionary Learning (DSDL) which provides a innovative solution towards multi-label image classification from the perspective of dictionary learning. DSDL is an end-to-end trainable deep learning model which incorporates the semantic information of categories and provides a naive but elegant way for exploiting and reconciling the label, semantic and visual spaces. The proposed model considers the multi-label image classification as a feature-dictionary representation problem in which the word embedding of categories is used for deriving semantic dictionary and the label embedding is deemed as the coefficients for representing the visual features of given sample. DSDL consists of three modules: feature learning, semantic dictionary generation and visual features representation modules. Figure~\ref{overview_pic} illustrates a detailed pipeline of DSDL, including semantic dictionary learning and multi-label image classification stages. Because representation coefficients and semantic dictionary cannot be solved directly like traditional dictionary learning, we design a novel Alternately Parameter Updating Strategy (APUS) for DSDL to realize the joint learning of the representation coefficients and semantic dictionary. APUS can optimize representation coefficients and semantic dictionary alternately during the forward and backward propagation of network via establishing their connections in dictionary learning.

\subsection{Deep Semantic Dictionary Learning}
\subsubsection{Feature Learning Module}
The CNN is currently the dominant feature learning approach. Here, we follow previous works~\cite{zhu2017learning, chen2019multi,chen2019learning} which adopt ResNet-101 pre-trained on ImageNet as the backbone for feature learning. Give an input image $x$, we feed it into ResNet-101 to obtain $2048\times 14 \times 14$ dimensional feature maps from the final convolution layer. Then, we apply global max-pooling to obtain the 2048-dimensional global feature $f$ as follows,
\begin{equation}
    f = \mathcal{F}(x) = f_{pool}(f_{cnn}(x; \theta)) \in \mathcal{R}^d \label{cnn},
\end{equation}
where $\theta$ indicates the learnable parameters of CNN and $d=2048$.

\subsubsection{Semantic Dictionary Generation Module}
The quality of semantic dictionary is the key to the success of our model. We intend to derive semantic dictionary in visual space from the word embedding of categories, since semantic information has been proven its discriminative characteristics in multi-label image classification~\cite{acm, aia, chen2019multi, chen2019learning, lin2015unsupervised}. In details, we employ an auto-encoder to accomplish the dictionary generation in nonlinear manner. Many NLP researches have demonstrated that auto-encoder can exploit the semantic spaces excellently and alleviate the overfitting of model by bidirectional transformation.

In our model, the GloVe~\cite{glove} trained on Wikipedia dataset is leveraged for extracting a $k$-dimensional word embedding for each class,
\begin{equation}
    s_i=f_{g}(w_i) \in \mathcal{R}^k \label{GloVe}
\end{equation}
where $w_i$ is the semantic word of $i$-th class and $s_i$ is the corresponding word embedding. Then, the encoder $f_{en}$ projects the word embeddings $S=[s_1,s_2, \dots ,s_c]\in \mathcal{R}^{k\times c}$ into the visual space to generate the semantic dictionary $D=[a_1, a_2, \dots ,a_c]\in \mathcal{R}^{d\times c}$ where $a_i \in \mathcal{R}^{d}$ is the $i$-th dictionary atom, and the decoder projects the semantic dictionary back to the semantic space for reconstructing $S$. Such procedure can be mathematically denoted as follows,
\begin{eqnarray}\label{encoder}
    D&=&f_{en}(S; \psi), \\  \label{decoder}
    \hat{S}&=&f_{de}(D;\psi^T),
\end{eqnarray}
where $f_{en}(\cdot)$ and $f_{de}(\cdot)$ share the same parameters. $\psi$ represents the learnable parameters. To ensure the learned semantic dictionary maintaining original semantic information as much as possible, the similarity between the original word embeddings $S$ and the reconstructed one $\hat{S}$ should be maximized, for which we adopt the cosine distance for measuring their discrepancy. The objective of auto-encoder is,
\begin{equation}
    \begin{aligned}
        \arg\underset{\psi}\max L_{sim}: & = \frac{1}{c} \sum_{i = 1}^{c}\cos(s_i,\hat{s_i})                                                        \\
                                         & = \frac{1}{c} \sum_{i = 1}^{c} \frac{s_i\cdot \hat{s_i}}{\|s_i\|_2 \times \|\hat{s_i}\|_2},  \label{sim}
    \end{aligned}
\end{equation}
where $\cos(\cdot)$ represents cosine function.

\subsubsection{Visual Features Representation Module}
In DSDL, we utilize the learned semantic dictionary to represent the visual features with the representation coefficients $\alpha=[\alpha^1,\alpha^2, \dots ,\alpha^c] \in \mathcal{R}^c$ whose normalized version $\hat{y}=\sigma(\alpha)=[\hat{y}^1,\hat{y}^2,\dots,\hat{y}^c]$ are deemed as the label occurrence probabilities with respect to visual features $f$, where $\sigma(\cdot)$ is the sigmoid function to normalize the representation coefficients. The reconstruction loss of the visual features is represented as
\begin{equation}
    L_{rec}=\|f-D \alpha\|_2^2, \label{rec}
\end{equation}
where $\alpha^i$ is the representation coefficient corresponding to the $i$-th atom $a_i$ in the dictionary.

DSDL is quite different from conventional dictionary learning approaches of which dictionaries are commonly overcomplete ($c>d$) while our dictionary is undercomplete ($c<d$). Thus, the proposed dictionary learning model should be considered as a collaborative representation rather than the conventional sparse representatio \cite{liu2019distributed}. The dictionary learning loss is updated as follows,
\begin{equation}
    L_{dic}=L_{rec}+\lambda \|\alpha\|_2^2 =\|f-D \alpha\|_2^2 + \lambda \|\alpha\|_2^2, \label{dic}
\end{equation}
where $\lambda>0$ is the hyper-parameter for reconciling reconstruction loss and $L_2$-norm regularization. The $L_2$-norm regularization can not only fully utilize all the atoms of dictionary for guaranteeing the unique solution of DSDL, but also introduce the smooth representation to alleviate the overfitting. Meanwhile, the discrepancy between label distribution and predicted occurrence probabilities of categories is minimized and measured by cross entropy,
\begin{equation}
    L_{ce}= -\sum_{i=1}^N \sum_{j = 1}^{c} y_i^j \log (\hat{y}_i^j) + (1- y_i^j) \log (1-\hat{y}_i^j).   \label{CE}
\end{equation}
After the minimization of $L_{ce}$, the optimal dictionary learning model is obtained,
\begin{equation}
    \mathcal{J}(f)\leftarrow\arg\underset{D,\alpha}\min ~~L_{ce}+\beta L_{dic}, \label{dlm}
\end{equation}
where $\beta>0$ is the hyper-parameter for balancing the losses.

\subsubsection{Overall Objective}
Equations~\ref{sim} and \ref{dlm} are integrated as Equation~\ref{total} for overall optimization objective of our DSDL.
\begin{equation}
    \mathcal{T(\cdot)}\leftarrow \arg\underset{\theta,\psi}\min~~L_{total}=\frac{L_{ce} + \beta L_{dic}}{L_{sim}} \label{total}
\end{equation}
With the above formula, the multi-label image classification is cast as deep dictionary learning for solution via learning the optimal $\theta$ and $\psi$. The proposed DSDL establishes the correlations among the label, semantic and visual spaces in perspective of dictionary learning. Specifically, it exploits the semantic and visual spaces with CNN and auto-encoder respectively, then the learned semantic dictionary and the visual features are collaboratively utilized to exploit the label space.

\subsection{Alternately Parameter Updating Strategy}
Our proposed DSDL cannot be optimized with the conventional backward propagation by minimizing $L_{total}$, because semantic dictionary $D$ is unreliable and representation coefficients $\alpha$ is unknown. We design the Alternately Parameter Updating Strategy (APUS) to optimize our model, which consists of representation coefficients updating and semantic dictionary updating stages executed alternately in the forward and backward propagation.

\subsubsection{Update Representation Coefficients \bm{$\alpha$}}
In the forward propagation, an unreliable semantic dictionary $D$ is generated by Equation~\ref{encoder}. Then semantic dictionary $D$ and visual features $f$ in Equation~\ref{dic} are fixed, thus the optimization problem of Equation \ref{dic} is transformed into a regularized least square problem for obtaining the representation coefficients $\alpha$. More specifically, let the derivative of the dictionary loss with respect to $\alpha$ be zero,
\begin{eqnarray}\nonumber
    \frac{\partial L_{dic}}{\partial \alpha} & =&\frac{\partial ( \Tr (D\alpha-f)^T(D\alpha-f)+\lambda \alpha^T\alpha) )}{\partial \alpha} \\
    &=&2D^TDe-2D^T f+2\lambda \alpha =0.  \label{der}
\end{eqnarray}
Then we can obtain its closed-form solution of $\alpha$,
\begin{equation}
    \alpha=(D^TD+\lambda I)^{-1}D^T f.   \label{alpha}
\end{equation}

\subsubsection{Update Semantic Dictionary \bm{$D$}} In the backward propagation, representation coefficients $\alpha$ is fixed to the value obtained in the previous step. The conventional backward propagation is used for optimizing the semantic dictionary $D$ and the feature learning module by minimizing $L_{total}$. Afterwards, the semantic dictionary $D$ and the visual features $f$ will be updated for next step.

Finally, these updating steps are alternately executed until convergence. The details of the model optimization procedure is summarized in Algorithm~\ref{apus}.
\begin{algorithm}[h]
    \caption{Alternately Parameter Updating} \label{apus}
    \KwIn{Training set $\mathcal{D} =\{(x_i,y_i) | 1\leqq i \leqq N \}$;\\
    \qquad ~~~~ Semantic words $\mathcal{W}=[w_1,w_2, \dots ,w_c]$;\\
    \qquad ~~~~ Iterative number $I_t$; Learning rate $\eta $;
    }
    \KwOut{Semantic dictionary $D$;\\
        \qquad ~~~~~~ Feature learning module $\mathcal{F}(\cdot);$
    }
    Initialize $\theta $ with pre-trained ResNet-101;\\
    Initialize $\psi $ randomly;\\
    \For{$u = 1,2,\cdots,I_t$}{
        \For{$i = 1,2,\cdots,N$}{
            \textbf{//\quad forward propagation}\\
            1.For each $x_i$, extract visual features $f_i$ with the feature learning module;\\
            2.For each $w_i$, obtain word embedding vector $s_i$ with the auto-encoder;\\
            3.Generate semantic dictionary $D$ by Equation~\ref{encoder};\\
            4.Compute coefficients $\alpha$ by Equation \ref{alpha};\\
            5.Obtain occurrence probabilities of categories $\hat{y_i}$ with sigmoid function;\\
            \textbf{//\quad backward propagation}\\
            7.Update network parameters $\theta $ and $\psi $ by minimizing $L_{total}$ in Equation~\ref{total};
        }
    }
    Return $D$, $\theta$ and $\psi$
\end{algorithm}

\subsection{Multi-label Image Classification}
Once the DSDL trained, a reliable semantic dictionary $D$ and the mapping function $\mathcal{F}(\cdot)$ of the feature learning module are obtained. Given a test image $x^\prime $, the visual feature $f^\prime$ is extracted with the feature learning module $\mathcal{F}(\cdot)$. Then the label occurrence probabilities of $x^\prime$ can be predicted by accomplishing a dictionary query task as follows,
\begin{eqnarray} \nonumber
    \hat{y^\prime} &=& \sigma(\hat{\alpha})= \sigma((D^TD+\lambda I)^{-1}D^T f^\prime )\\
    &=& \sigma((D^TD+\lambda I)^{-1}D^T \mathcal{F}(x^\prime )).
\end{eqnarray}
Conventionally, if the estimated probability is greater than 0.5, the image is positive in this category and vice versa.

\begin{table*}[ht]
    \centering
    \fontsize{25}{40}\selectfont
    \resizebox{\textwidth}{!}{\begin{tabular}{c|cccccccccccccccccccc|c}
            \hline
            Method                         & aero          & bike          & bird          & boat          & bottle        & bus           & car           & cat           & chair         & cow           & table         & dog           & horse         & motor         & person        & plant         & sheep         & sofa          & train         & tv            & mAP\cr
            \hline
            \hline
            HCP\cite{wei2015hcp}           & 98.6          & 97.1          & 98.0          & 95.6          & 75.3          & 94.7          & 95.8          & 97.3          & 73.1          & 90.2          & 80.0          & 97.3          & 96.1          & 94.9          & 96.3          & 78.3          & 94.7          & 76.2          & 97.9          & 91.5          & 90.9\cr
            ResNet-101\cite{he2016deep}    & 99.5          & 97.7          & 97.8          & 96.4          & 65.7          & 91.8          & 96.1          & 97.6          & 74.2          & 80.9          & 85.0          & \textbf{98.4} & 96.5          & 95.9          & 98.4          & 70.1          & 88.3          & 80.2          & 98.9          & 89.2          & 89.9\cr
            FeV+LV\cite{yang2016exploit}   & 97.9          & 97.0          & 96.6          & 94.6          & 73.6          & 93.9          & 96.5          & 95.5          & 73.7          & 90.3          & 82.8          & 95.4          & 97.7          & 95.9          & 98.6          & 77.6          & 88.7          & 78.0          & 98.3          & 89.0          & 90.6\cr
            RCP\cite{wang2016beyond}       & 99.3          & 97.6          & 98.0          & 96.4          & 79.3          & 93.8          & 96.6          & 97.1          & 78.0          & 88.7          & 87.1          & 97.1          & 96.3          & 95.4          & \textbf{99.1} & 82.1          & 93.6          & 82.2          & 98.4          & 92.8          & 92.5\cr
            CNN-RNN\cite{wang2016cnn}      & 96.7          & 83.1          & 94.2          & 92.8          & 61.2          & 82.1          & 89.1          & 94.2          & 64.2          & 83.6          & 70.0          & 92.4          & 91.7          & 84.2          & 93.7          & 59.8          & 93.2          & 75.3          & \textbf{99.7} & 78.6          & 84.0\cr
            RDAR\cite{wang2017multi}       & 98.6          & 97.4          & 96.3          & 96.2          & 75.2          & 92.4          & 96.5          & 97.1          & 76.5          & 92.0          & 87.7          & 96.8          & 97.5          & 93.8          & 98.5          & 81.6          & 93.7          & 82.8          & 98.6          & 89.3          & 91.9\cr
            RARL\cite{chen2018recurrent}   & 98.6          & 97.1          & 97.1          & 95.5          & 75.6          & 92.8          & 96.8          & 97.3          & 78.3          & 92.2          & 87.6          & 96.9          & 96.5          & 93.6          & 98.5          & 81.6          & 93.1          & 83.2          & 98.5          & 89.3          & 92.0\cr
            RMIC\cite{he2018reinforced}    & 97.1          & 91.3          & 94.2          & 57.1          & \textbf{86.7} & 90.7          & 93.1          & 63.3          & 83.3          & 76.4          & \textbf{92.8} & 94.4          & 91.6          & 95.1          & 92.3          & 59.7          & 86.0          & 69.5          & 96.4          & 79.0          & 84.5\cr
            RLSD\cite{zhang2018multilabel} & 96.4          & 92.7          & 93.8          & 94.1          & 71.2          & 92.5          & 94.2          & 95.7          & 74.3          & 90.0          & 74.2          & 95.4          & 96.2          & 92.1          & 97.9          & 66.9          & 93.5          & 73.7          & 97.5          & 87.6          & 88.5\cr
            DELTA\cite{yu2019delta}        & 98.2          & 95.1          & 95.8          & 95.7          & 71.6          & 91.2          & 94.5          & 95.9          & 79.4          & 92.5          & 85.6          & 96.7          & 96.8          & 93.7          & 97.8          & 77.7          & 95.0          & 81.9          & 99.0          & 87.9          & 91.1\cr
            ML-GCN\cite{chen2019multi}$^*$ & 99.8          & 98.2          & 98.0          & 98.3          & 80.2          & 94.6          & 97.4          & 97.7          & 82.9          & 93.1          & 87.6          & 97.8          & 97.9          & 95.2          & 98.9          & 83.3          & 95.4          & 81.2          & 98.8          & 93.7          & 93.5\cr
            CoP\cite{wen2020multilabel}    & \textbf{99.9} & 98.4          & 97.8          & \textbf{98.8} & 81.2          & 93.7          & 97.1          & \textbf{98.4} & 82.7          & 94.6          & 87.1          & 98.1          & 97.6          & \textbf{96.2} & 98.8          & 83.2          & \textbf{96.2} & 84.7          & 99.1          & 93.5          & 93.8\cr
            \hline
            \hline
            DSDL                           & 99.8          & \textbf{98.7} & \textbf{98.4} & 97.9          & 81.9          & \textbf{95.4} & \textbf{97.6} & 98.3          & \textbf{83.3} & \textbf{95.0} & 88.6          & 98.0          & \textbf{97.9} & 95.8          & 99.0          & \textbf{86.6} & 95.9          & \textbf{86.4} & 98.6          & \textbf{94.4} & \textbf{94.4}\cr\hline
        \end{tabular}}
    \caption{The performance comparison on the Pascal VOC 2007 dataset. The bold one indicates the best performance. The sign ``*" indicates the reproduced results via using the source codes provided by the original authors.}
    \label{voc2007results}
    \vspace{-0.2cm}
\end{table*}

\begin{table*}[ht]
    \centering
    \fontsize{25}{40}\selectfont
    \resizebox{\textwidth}{!}{\begin{tabular}{c|cccccccccccccccccccc|c}
            \hline
            Method                           & aero          & bike          & bird          & boat          & bottle        & bus           & car           & cat           & chair         & cow           & table         & dog           & horse         & motor         & person        & plant         & sheep         & sofa          & train         & tv            & mAP\cr
            \hline
            \hline
            VGG19+SVM\cite{simonyan2014very} & 99.1          & 88.7          & 95.7          & 93.9          & 73.1          & 92.1          & 84.8          & 97.7          & 79.1          & 90.7          & 83.2          & 97.3          & 96.2          & 94.3          & 96.9          & 63.4          & 93.2          & 74.6          & 97.3          & 87.9          & 89.0\cr
            HCP\cite{wei2015hcp}             & 99.1          & 92.8          & 97.4          & 94.4          & 79.9          & 93.6          & 89.8          & 98.2          & 78.2          & \textbf{94.9} & 79.8          & 97.8          & 97.0          & 93.8          & 96.4          & 74.3          & 94.7          & 71.9          & 96.7          & 88.6          & 90.5\cr
            FeV+LV\cite{yang2016exploit}     & 98.4          & 92.8          & 93.4          & 90.7          & 74.9          & 93.2          & 90.2          & 96.1          & 78.2          & 89.8          & 80.6          & 95.7          & 96.1          & 95.3          & 97.5          & 73.1          & 91.2          & 75.4          & 97.0          & 88.2          & 89.4\cr
            RCP\cite{wang2016beyond}         & 99.3          & 92.2          & 97.5          & 94.9          & 82.3          & 94.1          & 92.4          & 98.5          & 83.8          & 93.5          & 83.1          & 98.1          & 97.3          & \textbf{96.0} & \textbf{98.8} & 77.7          & 95.1          & 79.4          & 97.7          & 92.4          & 92.2\cr
            RMIC\cite{he2018reinforced}      & 98.0          & 85.5          & 92.6          & 88.7          & 64.0          & 86.8          & 82.0          & 94.9          & 72.7          & 83.1          & 73.4          & 95.2          & 91.7          & 90.8          & 95.5          & 58.3          & 87.6          & 70.6          & 93.8          & 83.0          & 84.4\cr
            DELTA\cite{yu2019delta}          & -             & -             & -             & -             & -             & -             & -             & -             & -             & -             & -             & -             & -             & -             & -             & -             & -             & -             & -             & -             & 90.3\cr
            \hline
            \hline
            DSDL                             & \textbf{99.4} & \textbf{95.3} & \textbf{97.6} & \textbf{95.7} & \textbf{83.5} & \textbf{94.8} & \textbf{93.9} & \textbf{98.5} & \textbf{85.7} & 94.5          & \textbf{83.8} & \textbf{98.4} & \textbf{97.7} & 95.9          & 98.5          & \textbf{80.6} & \textbf{95.7} & \textbf{82.3} & \textbf{98.2} & \textbf{93.2} & \textbf{93.2} \cr\hline
        \end{tabular}}
        
    \caption{The performance comparison on the Pascal VOC 2012 dataset. The bold one indicates the best performance. The sign ``-" denotes the corresponding result is not provided.}
     \vspace{-0.2cm}
    \label{voc2012results}

\end{table*}
\begin{table*}[ht]
    \centering
    \fontsize{8}{12}\selectfont
    \begin{tabular}{c|c|cccccc|cccccc}
        \hline
        \multirow{2}{*}{Method}            &
                                           & \multicolumn{6}{c|}{TOP3} & \multicolumn{6}{c}{ALL}\cr\cline{2-14}
                                           & mAP                       & CP                                     & CR            & CF1           & OP            & OR            & OF1           & CP            & CR            & CF1           & OP            & OR            & OF1 \cr
        \hline
        \hline
        WARP\cite{gong2013deep}            & -                         & 59.3                                   & 52.5          & 55.7          & 59.8          & 61.4          & 60.7          & -             & -             & -             & -             & -             & -\cr
        CNN-RNN\cite{wang2016cnn}          & -                         & 66.0                                   & 55.6          & 60.4          & 69.2          & \textbf{66.4} & 67.8          & -             & -             & -             & -             & -             & -\cr
        ResNet-101\cite{he2016deep}        & 77.3                      & 84.1                                   & 59.4          & 69.7          & 89.1          & 62.8          & 73.6          & 80.2          & 66.7          & 72.8          & 83.9          & 70.8          & 76.8 \cr
        RDAR\cite{wang2017multi}           & 73.4                      & 79.1                                   & 58.7          & 67.4          & 84.0          & 63.0          & 72.0          & -             & -             & -             & -             & -             & -\cr
        ResNet-SRN\cite{zhu2017learning}   & 77.1                      & 85.2                                   & 58.8          & 67.4          & 87.4          & 62.5          & 72.9          & 81.6          & 65.4          & 71.2          & 82.7          & 69.9          & 75.8\cr
        RLSD\cite{zhang2018multilabel}     & -                         & 67.6                                   & 57.2          & 62.0          & 70.1          & 63.4          & 66.5          & -             & -             & -             & -             & -             & -\cr
        KD-WSD\cite{liu2018multi}          & 74.6                      & -                                      & -             & 66.8          & -             & -             & 72.7          & -             & -             & 69.2          & -             & -             & 74.0\cr
        RARL\cite{chen2018recurrent}       & -                         & 78.8                                   & 57.2          & 66.2          & 84.0          & 61.6          & 71.1          & -             & -             & -             & -             & -             & -\cr
        DELTA\cite{yu2019delta}            & 71.3                      & -                                      & -             & -             & -             & -             & -             & -             & -             & -             & -             & -             & -\cr
        ResNet101-ACfs\cite{guo2019visual} & 77.5                      & 85.2                                   & 59.4          & 68.0          & 86.6          & 63.3          & 73.1          & 77.4          & 68.3          & 72.2          & 79.8          & 73.1          & 76.3\cr
        CoP\cite{wen2020multilabel}        & 81.1                      & 86.4                                   & 62.9          & 72.7          & 88.7          & 65.1          & 75.1          & 81.2          & \textbf{70.8} & 75.8          & 83.6          & 73.3          & 78.1\cr
        \hline
        \hline
        DSDL                               & \textbf{81.7}             & \textbf{88.1}                          & \textbf{62.9} & \textbf{73.4} & \textbf{89.6} & 65.3          & \textbf{75.6} & \textbf{84.1} & 70.4          & \textbf{76.7} & \textbf{85.1} & \textbf{73.9} & \textbf{79.1}\cr\hline
    \end{tabular}
    \caption{The performance comparison on the Microsoft COCO dataset. The bold one indicates the best performance. The sign ``-" denotes the corresponding result is not provided.}
    \label{cocoresults}
     \vspace{-0.2cm}
\end{table*}

\section{Experiments}
\subsection{Experimental Settings}
\subsubsection{Datasets:} To prove the effectiveness of DSDL, we conduct extensive experiments on three public multi-label image benchmarks are used for model evaluation, \textit{i.e.}, Pascal VOC 2007, Pascal VOC 2012 and Microsoft COCO.

\textit{Pascal VOC 2007}~\cite{voc} is the most widely used datasets to evaluate the multi-label image classification task, which contains 20 categories divided into training (2,501), validation (2,510) and testing (4,952). Following the previous work, we train our model on training and validation sets, and evaluate on the testing set.

\textit{Pascal VOC 2012}~\cite{voc} consists of the same 20 categories as VOC 2007 while VOC 2012 contains images from 2008-2011 and there is no intersection with 2007. VOC 2012 dataset is divided into training (5,717), validation (5,823) and testing (10,991). We train our model on training set, and fine-tune on validation set. Since there are no ground truth labels provided for the testing set, all approaches are evaluated by submitting the testing results to the Pascal VOC Evaluation Server.

\textit{Microsoft COCO}~\cite{lin2014microsoft} is large scale images with Common Objects in Context (COCO), which is widely applied to many fields such as object detection, segmentation, and image caption. COCO contains 122,218 images and covers 80 common categories, which is further divided into 82,081 images training set and 40,137 images validation set. We evaluate the performance of all the methods on the validation set.

\subsubsection{Evaluation Metrics:}We follow the conventional setting \cite{chen2019multi} which adopts the average precision (AP) on each category and mean average precision (mAP) over all categories for evaluation. With regard to the COCO dataset, we assign the labels with top-3 scores for each image and compare them with the ground truth labels. More specifically, the overall precision, recall, F1-measure (OP, OR, OF1) and per-class precision, recall, F1-measure (CP, CR, CF1) are employed as detailed performance indicators,
\begin{equation}
    \begin{split}
        OP&=\frac{\sum_i N_i^t}{\sum_i N_i^p}, \qquad CP=\frac{1}{c} \sum_i \frac{N_i^t}{N_i^p}, \\
        OR&=\frac{\sum_i N_i^t}{\sum_i N_i^g}, \qquad CR=\frac{1}{c} \sum_i \frac{N_i^t}{N_i^g}, \\
        OF1&=\frac{2\times OP \times OR}{OP + OR}, \quad CF1=\frac{2\times CP \times CR}{CP + CR},\\
    \end{split}
\end{equation}
where $c$ is the number of labels, $N_i^t$ is the number of images that are correctly predicted for the $i$-th label, $N_i^p$ is the number of predicted images for the $i$-th label, $N_i^g$ is the number of ground truth images for the $i$-th label. Generally speaking, OF1, CF1 and mAP are the relatively more important performance indicator.

\subsubsection{Compared Methods:} We compare our proposed DSDL with dozens of multi-label image classification approaches. Most of them are based on deep learning. They are VGGNet+SVM~\cite{simonyan2014very}, HCP~\cite{wei2015hcp}, FeV+LV~\cite{yang2016exploit}, RCP~\cite{wang2016beyond}, RMIC~\cite{he2018reinforced},  WARP~\cite{gong2013deep}, CNN-RNN~\cite{wang2016cnn}, ResNet-101~\cite{he2016deep}, RDAR~\cite{wang2017multi}, ResNet-SRN-att~\cite{zhu2017learning}, RLSD~\cite{zhang2018multilabel}, KD-WSD~\cite{liu2018multi}, RARL~\cite{chen2018recurrent}, DELTA~\cite{yu2019delta}, ResNet101-ACfs~\cite{guo2019visual}, ML-GCN~\cite{chen2019multi} and CoP~\cite{wen2020multilabel}.

\subsubsection{Implementation Details:}
ResNet-101 pre-trained on ImageNet is utilized as the feature learning module. The input images are randomly cropped and resized into $448 \times 448$ with random horizontal flips for data augmentation. With regard to the dictionary learning module, the encoder consists of two fully connected layers with output dimension of 1024 and 2048 followed by LeakyReLU with negative slope 0.2. The decoder shares the same learnable parameters with encoder. All modules are optimized with Stochastic Gradient Descent (SGD) optimizer. The momentum is 0.9 and the weight decay is $10^{-4}$. The initial learning rate is 0.01, which decays by a factor of 10 for every 40 epochs and the network is trained for 100 epochs in total.

\subsection{Experimental Results}
Table~\ref{voc2007results}, Table~\ref{voc2012results} and Table~\ref{cocoresults} tabulate the performances of different multi-label image classification methods on VOC 2007, 2012 and COCO datasets respectively. The results show that DSDL achieves the best performances in mAP on all three datasets where mAP is deemed as the most important comprehensive performance indicator in multi-label image classification. More specifically, DSDL shows its advantage almost in all the object recognition tasks on two VOC datasets. The performance gains of DSDL over DELTA, ML-GCN and CoP, which are the three of the most recent compared approaches on VOC 2007 dataset, are 3.3\%, 0.9\% and 0.6\% respectively, where ML-GCN and CoP also achieved the ResNet101 as their backbone networks. DSDL performs more dominant on VOC 2012 dataset where it possesses 1.0\% more accuracies than the second performed method, and only fails to win the first on the recognition of cow, motor and person with a narrow margin.

On COCO dataset, DSDL achieves the best mAP, CF1 and OF1 which are all the most important performance indicators. Its CF1 and OF1 in TOP3 case are 73.4\% and 75.6\%. These numbers in ALL case are 76.7\% and 79.1\%. Compared with the plain RestNet101 model, DSDL gets 4.4\% more accuracies in mAP which validates the effectiveness of our proposed DSDL. Moreover, DSDL consistently performs better than the other ResNet101-based approaches, such as ResNet-SRN, ResNet101-ACfs and CoP. The gains over these methods in mAP are 4.6\%, 4.2\% and 0.6\% respectively on COCO dataset. Overall, although DSDL is succinct which just introduces the idea of dictionary learning to multi-label image classification for reconciling the all three involved spaces, all the experimental results imply that it achieves promising performance and has potential for further improvement.

\begin{figure}[t]
    \centering
    \subfigure[Pascal VOC]{
        \includegraphics[scale=0.26]{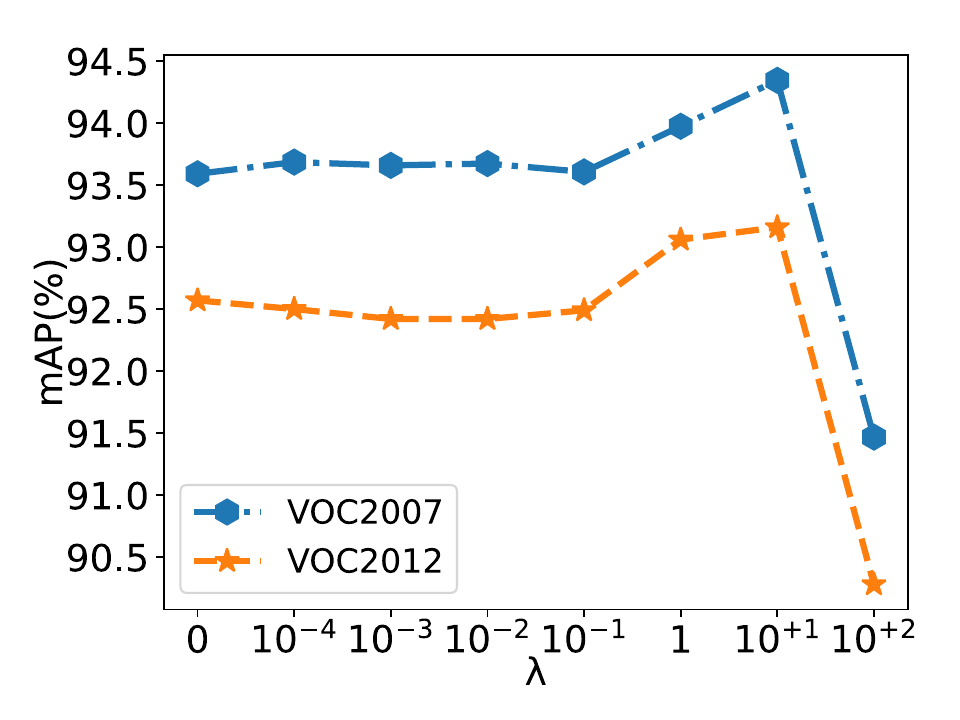}
    }
    \subfigure[MS COCO]{
        \includegraphics[scale=0.26]{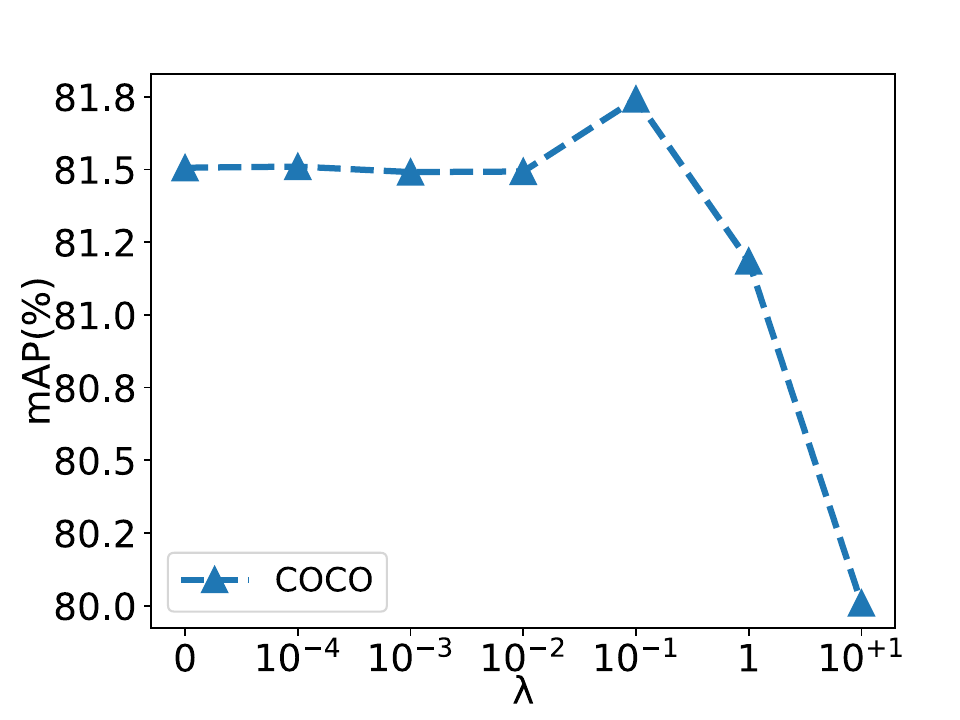}
    }
    \caption{Performance comparisons with different values of $\lambda$. Note that, when $\lambda = 0$, there is no regularizer.}
    \label{ablation_lambda}
\end{figure}

\begin{figure}[t]
    \centering
    \subfigure[Pascal VOC]{
        \includegraphics[scale=0.26]{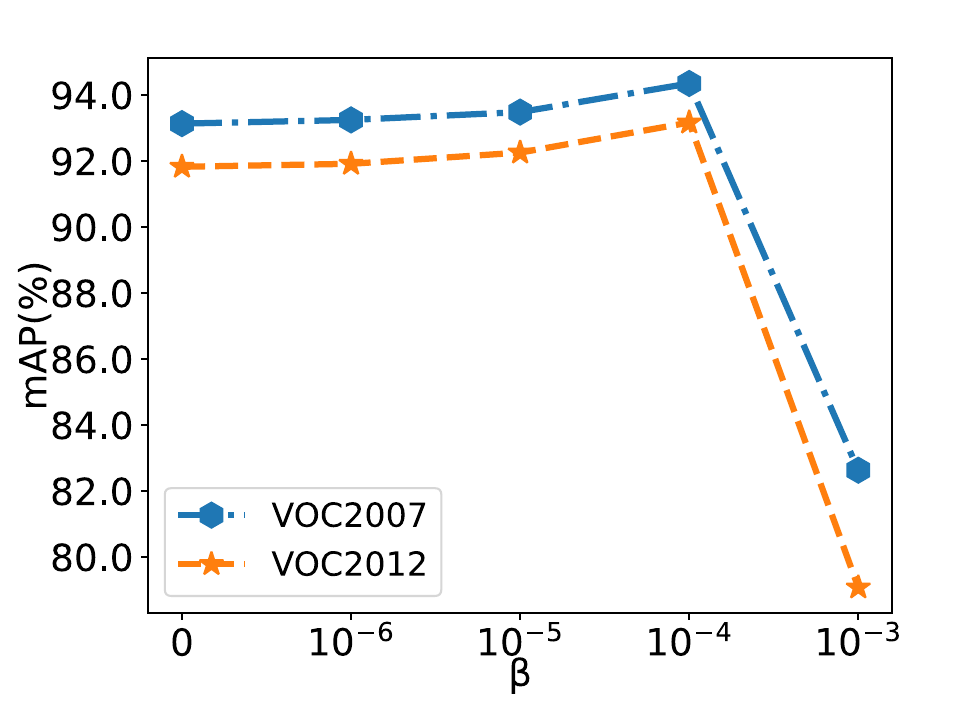}
    }
    \subfigure[MS COCO]{
        \includegraphics[scale=0.26]{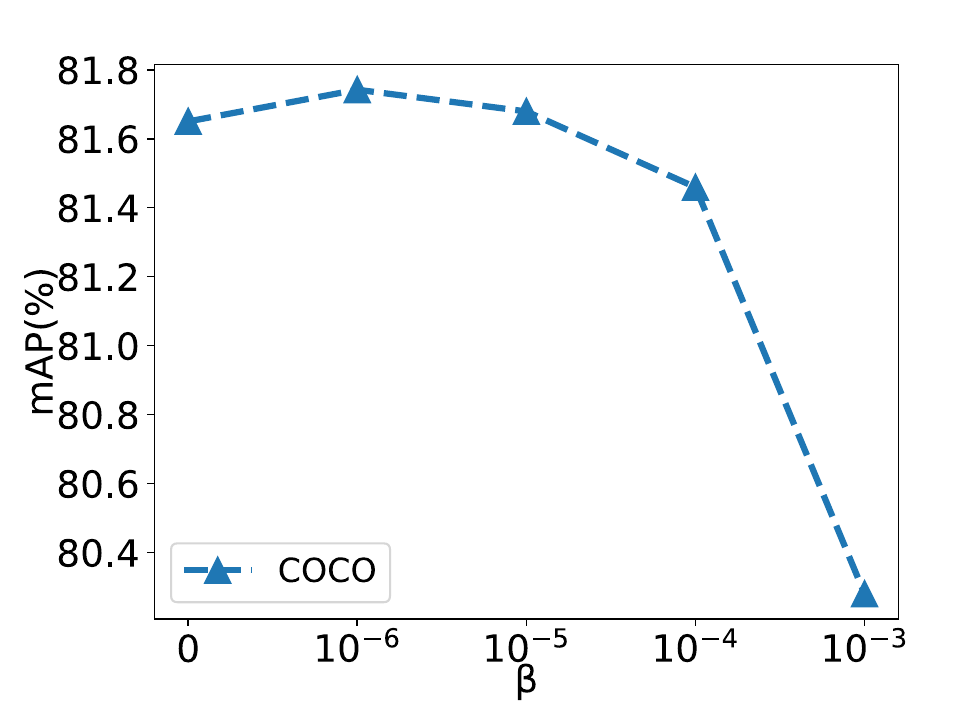}
    }
    \caption{Performance comparisons with different values of $\beta$. Note that, when $\beta = 0$, dictionary learning loss $L_{dic}$ is ignored.}
    \label{ablation_beta}
\end{figure}
\subsection{Ablation Studies}
\subsubsection{Effects of different hyper-parameters}
There are two manually tunable hyper-parameters $\lambda$ and $\beta$ which are used for controlling the impacts of $L_2$-norm regularization and dictionary learning loss respectively.
We conduct ablation experiments on three datasets to analyze the impacts of these parameters. Figures \ref{ablation_lambda} and \ref{ablation_beta} report their experimental results. It can be observed that DSDL seems to be more sensitive to $\lambda$ than $\beta$. A greater $\lambda$ often leads to a better result until $\lambda$ is too greater to lead to the model collapse while $\beta$ have the similar phenomenon but the performance is more stable when its value is below $10^{-4}$. Another interesting phenomenon is the performance improvement is not so significant when the $\beta$ is set from zero to $10^{-4}$. We attribute this to the fact that DSDL is still benefited from the dictionary learning even without the dictionary losses due to the specialty of Alternately Parameter Updating Strategy (APUS) which performs the dictionary learning in forward propagation and then aligns the normalized representation coefficients as the predicted labels with ground truths in the backward propagation alternately. In other words, the idea of dictionary learning has been already integrated into DSDL essentially in the parameter updating stage and the dictionary learning loss is just an additional emphasis on the dictionary representation of a sample in backward propagation step. With regard to the COCO dataset, the performances under different $\beta$ and $\lambda$ are more stable but they still suffer from the model collapse when their values keep increasing. In summary, we suggest $\lambda=10$, $\beta =10^{-4}$ for Pascal VOC and  $\lambda=0.1$, $\beta =10^{-6}$ for COCO.

\section{Conclusion}
In this paper, we address the problem of multi-label image classification by considering it as a dictionary learning task and propose a novel end-to-end approach, Deep Semantic Dictionary Learning (DSDL). With adopting dictionary learning technique, the DSDL exploits and reconciles all involved spaces, including label, semantic and visual spaces, through generating the semantic dictionary and taking the reconstruction of visual features as a dictionary query task to obtain normalized representation coefficients as label occurrence probabilities. Our proposed DSDL approach is only a plain version of above idea without applying any extra tricks like the attention mechanism or label correlation analysis, while extensive experiments on three standard multi-label image benchmarks shown that it does achieved promising performances in comparison with the state-of-the-arts. In future study, we will consider introducing attention mechanism or label correlation to further improve the performance of our model.

\section{Acknowledgements}
This work was supported in part by the National Natural Science Foundations of China under Grant 61602068, the Fundamental Research Funds for the Central Universities under Grant 2019CDCGRJ314 and 2019CDYGYB014.

\bibliography{references}

\end{document}